\documentclass{MIPRO}

\usepackage{cite}
\usepackage{amsmath,amssymb,amsfonts}
\usepackage{algorithmic}
\usepackage{graphicx}
\usepackage{textcomp}
\usepackage{xcolor}
\usepackage[T1]{fontenc}
\usepackage{flushend}
\usepackage{multirow}
\usepackage{multicol}
\usepackage{array}
\usepackage{url}
\usepackage{comment}

\begin{document}

\title{Market-Based Replanning for Safety-Critical UAV Swarms in Search and Rescue Missions}

\author{
\IEEEauthorblockN{
Luiz Giacomossi,
Andrea Haglund,
Claire Namatovu,
Emily Zainali,
Esaias Målqvist,\\
Yonatan M. Beyene,
Ivan Tomasic,
Baran C\c{u}r\"{u}kl\"{u}, and
Håkan Forsberg
}
\IEEEauthorblockA{Mälardalen University (MDU)\\
Västerås, Sweden\\
luiz.giacomossi@mdh.se
}

}

\maketitle


\begin{abstract}
Reliable autonomous UAV swarms in Search and Rescue (SAR) missions require fault-tolerant coordination capable of sustaining operations despite agent degradation. This paper introduces the Intelligent Replanning Drone Swarm, a distributed coordination architecture designed for resource-constrained environments. The proposed framework employs a \textit{Reverse-Auction} market mechanism where agents bid to service search sectors based on a distance-weighted cost function, coupled with a geometric consensus protocol for target verification. We evaluate the approach through physics-based simulations ($N=8$ agents, $8 \times 8$ grid) subjected to stochastic fault injection. Results indicate that the swarm autonomously reallocates tasks from failed agents with low latency relative to the total mission duration, maintaining a mission success rate of $93\%$ under $25\%$ workforce degradation. The proposed framework demonstrates a robust, empirically tested method for self-healing aerial robotic coordination.
\end{abstract}

\renewcommand\IEEEkeywordsname{Keywords}


\section{Introduction}
\label{sec:introduction}

Autonomous Unmanned Aerial Vehicle (UAV) swarms are increasingly utilized for Search and Rescue (SAR) \cite{Search_MDU,LARS_Search}, defense operations \cite{loyalwingman}, and disaster response due to their potential for redundancy and parallel task execution \cite{brambilla2013swarm}. However, deploying these systems in uncontrolled real-world scenarios presents reliability challenges. A primary requirement is \textit{Fault Tolerance}; specifically, the ability to sustain mission-critical functions despite individual agent failures (termed \textit{Fail-Operational} at the mission level).

Centralized control architectures create a single point of failure and are limited by communication latency in large-scale deployments. Conversely, distributed approaches may lack the predictable safety properties required for critical operations. This paper addresses the problem of maintaining swarm integrity and mission continuity during agent degradation (e.g., battery depletion or crash).

We present \textit{Intelligent Replanning Drone Swarm} (IRDS), a decentralized framework for resilient task allocation designed for resource-constrained environments (low-bandwidth, limited-compute and energy). The proposed method integrates a \textit{Reverse-Auction Market}—where agents minimize the collective cost of service—with a spatial consensus protocol for target verification. This enables the swarm to detect node failures, modeled as "Liquidation events" (immediate task release), update the global system state, and redistribute search sectors to maintain coverage.

Our contributions are:
\begin{enumerate}
    \item \textbf{Reverse-Auction Protocol:} We formalize a decentralized task allocation mechanism using distance-weighted pricing to minimize collective travel distance and improve target coverage.
    \item \textbf{Resilient Architecture:} We propose a fault-tolerant state machine that treats agent failure as a market event, triggering automatic task reallocation without central intervention.
    \item \textbf{Empirical Validation:} We validate robustness through physics-based simulations (PyBullet) with stochastic fault injection. Results demonstrate that the framework maintains mission success rates of $93\%$ even under significant workforce loss.
\end{enumerate}


The paper is organized as follows: Section~\ref{sec:related_work} reviews related work; Section~\ref{sec:problem_formulation} formalizes the problem; Section~\ref{sec:method} details the IRDS architecture; Sections~\ref{sec:experiments} and~\ref{sec:discussion} present and discuss results; Section~\ref{sec:conclusion} concludes.

\section{Related Work}
\label{sec:related_work}

Current research in multi-agent swarm coordination can be classified into centralized optimization, heuristic search strategies, and market-based approaches.

\subsection{Heuristic and Probabilistic Search}
Coordinate search is often achieved through shared world models or emergent behaviors. Probabilistic approaches typically integrate Bayesian updates with local gradients to guide agents toward high-entropy areas. For instance, Finite State Machines (FSM) coupled with potential fields have been shown to effectively coordinate collision-free swarms \cite{LARS_Search}. More advanced methods utilize probabilistic priors to reduce search times compared to blind lawn-mower patterns \cite{Search_MDU}. However, these heuristic approaches typically depend on continuous state synchronization and often lack built-in fault tolerance to automatically redistribute tasks when an agent fails.

\subsection{Task Allocation: Optimization vs. Markets}
The allocation of agents to tasks is a known problem in swarm robotics, often formulated as the Multi-Robot Task Allocation (MRTA) problem \cite{khamis2015multi}. In defense contexts, global optimization techniques such as the \textit{Hungarian Algorithm} (Kuhn-Munkres) can guarantee optimal assignment of assets to threats \cite{loyalwingman}. However, global optimization typically entails a computational complexity of $O(N^3)$, which scales poorly for large swarms and creates a centralized bottleneck.

Market-based approaches offer a decentralized alternative by employing virtual economies where agents bid for tasks based on local cost functions \cite{gerkey2002sold, khamis2015multi}. By prioritizing local cost minimization, auction mechanisms reduce algorithmic complexity to approximately $O(N \log N)$. This enables the high-frequency replanning required for dynamic SAR environments, avoiding the computational penalty of combinatorial optimization \cite{athira2024systematic}. IRDS adopts this paradigm but simplifies the bidding process to distance-weighted metrics to minimize communication overhead.

\vspace{-2mm}

\subsection{Consensus in Ad-Hoc Networks}
Reliable target verification in swarms requires distributed consensus \cite{amirkhani2022consensus} to filter false positives. While blockchain and Paxos algorithms offer cryptographically secure consistency, they impose bandwidth overheads often untenable for low-power drone meshes \cite{choi2009consensus}. Consequently, IRDS utilizes a lightweight geometric voting protocol that decouples exploration from verification, ensuring that sensor noise does not derail the global mission.

\section{Problem Formulation \& Mathematical Model}
\label{sec:problem_formulation}

We consider a coordination of $N$ autonomous agents, $\mathcal{D} = \{d_1, \dots, d_N\}$, to search a bounded area $\mathcal{A} \subset \mathbb{R}^2$ discretized into $M$ non-overlapping sectors $\mathcal{S} = \{s_1, \dots, s_M\}$. The state of agent $i$ at time $t$ is $\mathbf{x}_i(t) = [\mathbf{p}_i(t), b_i(t)]^T$, representing position and battery level. The objective is to maximize the covered area while minimizing collective travel distance, subject to dynamic constraints imposed by agent failures and communication limits.

\vspace{-2mm}
\subsection{Cost Function Definition}
To quantify travel overhead of task execution, we define a \textit{Service Cost} function based on proximity. The bid price $B_{ij}(t)$ for agent $d_i$ to acquire sector $s_j$ is modeled as:
\begin{equation}
    B_{ij}(t) = C_{base} \cdot \left( 1 + \frac{\| \mathbf{p}_i(t) - \mathbf{p}_{s_j} \|}{\delta} \right)
    \label{eq:bid_cost}
\end{equation}
\noindent where $C_{base}$ is the nominal base cost, $\mathbf{p}_{s_j}$ is the sector centroid, and $\delta$ is the distance scaling factor. This serves as the basis for the reverse-auction protocol (Sec.\ref{sec:method}).

\subsection{Dynamic Consensus Condition}
Target verification is modeled as a health-aware spatial voting problem. Let $\mathcal{V}(t) \subseteq \mathcal{D}$ be the set of currently active recruited verifiers (neighboring agents temporarily assigned to confirm a detected target). The set is updated dynamically based on agent health $h_i(t)$:
\begin{equation}
    \mathcal{V}(t) = \{d_v \in \mathcal{V}_{init} \mid h_v(t) = \text{OK}\}
\end{equation}
A target is confirmed if and only if the number of positive confirmations $N_{pos}$ satisfies the dynamic unanimity rule:
\begin{equation}
    N_{pos}(t) = |\mathcal{V}(t)| \quad \text{where} \quad |\mathcal{V}(t)| > 0
    \label{eq:dynamic_quorum}
\end{equation}
This mechanism removes crashed verifiers from the quorum during voting, preventing consensus deadlock caused by agent unavailability.

\vspace{-1mm}
\subsection{Distributed Collision Avoidance}
Safety is enforced via a decentralized, modified heuristic Artificial Potential Field (APF) \cite{khatib1986real}. The repulsive force $\mathbf{F}_{ij}$ acting on agent $d_i$ from neighbor $d_j$ applies a squared inverse-distance barrier to penalize proximity:
\begin{equation}
    \mathbf{F}_{ij} = 
    \begin{cases} 
        k_{rep} \left( \frac{1}{d_{ij}} - \frac{1}{r_{safe}} \right)^2 \frac{\mathbf{p}_{ij}}{d_{ij}} & \text{if } d_{ij} < r_{safe} \\
        \mathbf{0} & \text{otherwise}
    \end{cases}
    \label{eq:apf}
\end{equation}
\noindent where $d_{ij} = \| \mathbf{p}_i - \mathbf{p}_j \|$ is the Euclidean distance, $\mathbf{p}_{ij} = \mathbf{p}_i - \mathbf{p}_j$ is the relative position vector, $k_{rep}$ is the repulsion gain, and $r_{safe}$ is the safety radius interaction horizon.

\vspace{-2mm}

\section{Proposed Method: The IRDS Architecture}
\label{sec:method}

\begin{figure*}[ht]
    \centering
    \includegraphics[width=0.9\textwidth]{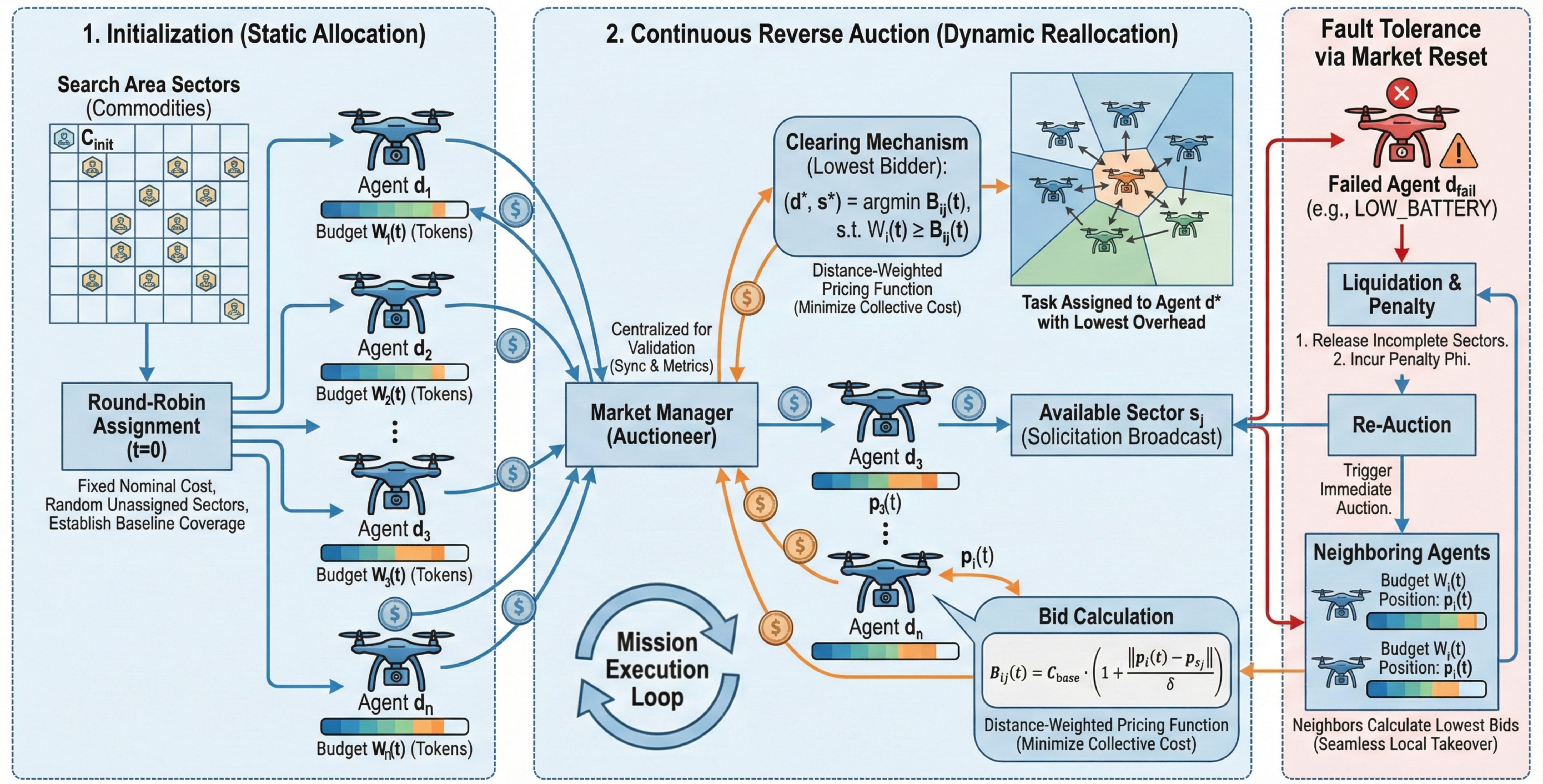} 
    \vspace{-2mm}
    \caption{Architectural overview of the IRDS Token-Based Reverse Auction. The system operates in three phases: (1) Static initialization; (2) Dynamic reallocation where agents minimize service cost; and (3) Fault recovery via market liquidation.} 
    \label{fig:architecture}
\end{figure*}

The IRDS architecture employs a hierarchical control strategy decomposing the mission into three layers: Global Task Allocation, Local Trajectory Generation, and Reactive Control. The system architecture is visualized in Fig. \ref{fig:architecture}.

\vspace{-1mm}
\subsection{Layer 1: Market-Based Task Allocation}
The search space $\mathcal{S}$ is treated as a set of discrete commodities. The allocation process is divided into two distinct phases.

\subsubsection{Phase I: Static Initialization}
At $t=0$, to prevent initial clustering, the system executes a round-robin assignment. All sectors are assigned a fixed nominal cost $C_{base}$. Agents iteratively acquire an initial set of unassigned sectors to ensure a uniform spatial distribution across the search area.

\subsubsection{Phase II: Dynamic Reverse Auction}
During mission execution, task allocation shifts to a distance-weighted reverse auction. When a sector becomes available (i.e., is unassigned or released via liquidation), valid bids $B_{ij}$ are established via the cost function in Eq. (\ref{eq:bid_cost}). The allocation logic clears the market by identifying the lowest global bidder:
\begin{equation}
    (d^*, s^*) = \arg\min_{d_i \in \mathcal{D}} B_{ij}(t) \quad \text{s.t.} \quad W_i(t) \ge B_{ij}(t)
\end{equation}

\subsubsection{Fault Tolerance (Liquidation)}
Upon detection of a fault (e.g., \texttt{BAD\_BATTERY}), the system executes a Liquidation Protocol: (1) All sectors owned by the faulty agent are released; (2) These sectors are re-auctioned immediately. Neighbors naturally win these auctions due to proximity, facilitating rapid autonomous takeover.

\subsection{Layer 2: Coverage \& Trajectory}
Upon securing a task, the agent employs a deterministic \textit{Boustrophedon Decomposition} \cite{cabreira2019survey}. This decomposition arranges trajectories in a ``lawnmower'' pattern to guarantee complete coverage of the assigned region. The planner generates a reference trajectory $\mathcal{P}_j$ relative to the section centroid, with sweep spacing determined by the sensor's Field of View (FOV). Simultaneously, the agent continuously computes the repulsive forces $\mathbf{F}_{ij}$ defined in Eq. (\ref{eq:apf}) to modify the control input for real-time collision avoidance.

\begin{figure}[ht]
    \centering
    \includegraphics[width=\columnwidth]{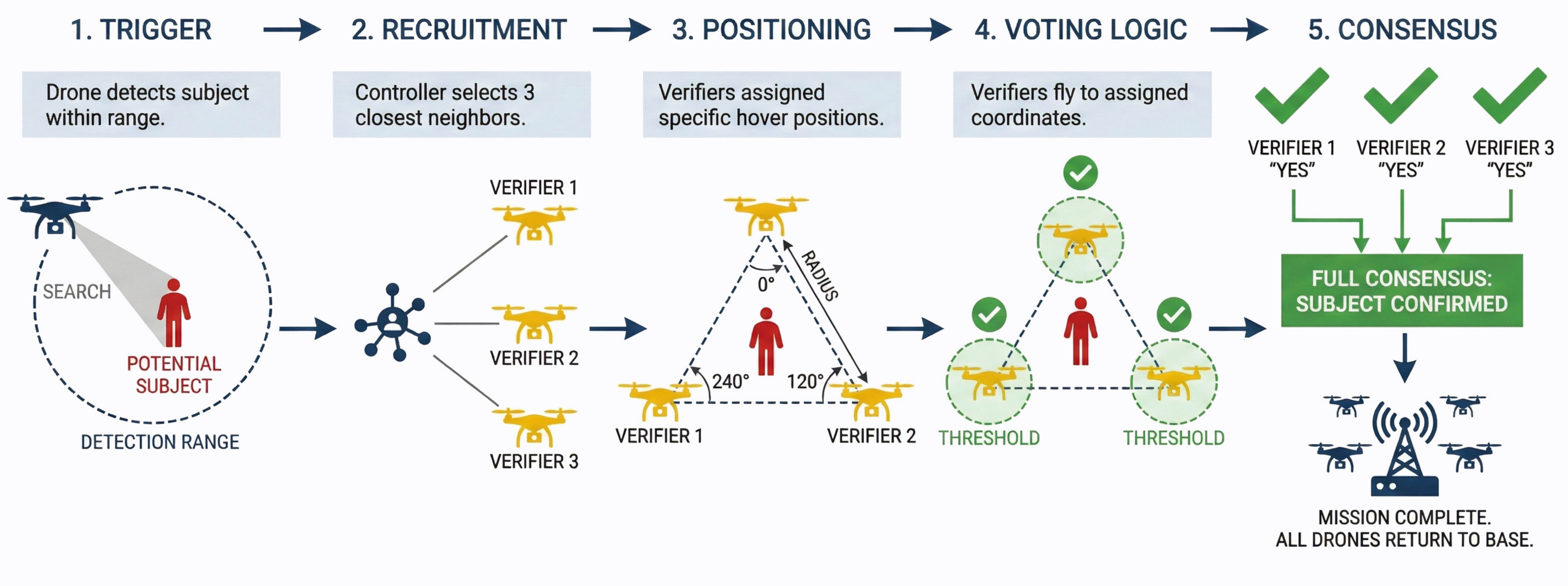} 
    \vspace{-8mm}
    \caption{Visual representation of the Multi-Agent Consensus Protocol. (1) Trigger: Detector identifies target. (2) Recruitment: The $k=3$ nearest neighbors are tasked as verifiers. (3) Positioning: Verifiers form an equilateral triangle (geometric lock) around the target. (4) Consensus: Unanimity is required to confirm the target and terminate the mission.}
    \label{fig:voting}
\end{figure}

\vspace{-2mm}

\subsection{Layer 3: Geometric Consensus Protocol}
To validate target detection in noisy environments, IRDS implements the multi-stage voting process illustrated in Fig. \ref{fig:voting}.
\begin{enumerate}
    
    \item \textbf{Recruitment:} Upon initial detection, the detecting agent $d_{det}$ halts its search and recruits the $k=3$ nearest neighbors to establish the candidate set $\mathcal{V}_{init}$.
    
    \item \textbf{Geometric Positioning:} To ensure diverse vantage points, the verifiers are commanded to form an \textit{Equilateral Triangle} formation (in the nominal $k=3$ case) around the estimated target coordinates (0\textdegree, 120\textdegree, 240\textdegree).
    
    \item \textbf{Voting Logic:} Once in position, each verifier casts a binary vote based on its local sensor reading.
    
    \item \textbf{Consensus:} The mission concludes only if the dynamic unanimity condition ($N_{pos} = |\mathcal{V}(t)|$) is met. Agent failures during this phase automatically reduce $|\mathcal{V}(t)|$ (Eq. \ref{eq:dynamic_quorum}) to prevent deadlocks. If the condition is subsequently not met (i.e., healthy voters disagree), the target is discarded as a false positive and verifiers return to the market.
\end{enumerate}

\vspace{-2mm}
\subsection{Implementation Note}
\label{sec:implementation}
The experiments utilize a central \texttt{MarketSystem} class within the simulation loop to manage currency and synchronization, acting as a proxy for a distributed ledger. The bidding logic, liquidation protocol, and consensus voting are computed locally by each agent and are decentralized by construction (only arbitration and metric collection are centralized in simulation). In a physical network, the auctioneer role maps to a distributed state machine synchronized via ad-hoc gossip protocols. Thus, real-world RF latency and packet loss can increase reallocation delays and introduce inconsistencies in the global belief state.

\vspace{-1mm}

\section{Experimental Evaluation}
\label{sec:experiments}

To validate the dependability and efficiency of the IRDS architecture, we conducted extensive simulations in a high-fidelity physics environment.
\vspace{-1mm}

\subsection{Operational Scenario Description}
\begin{figure*}[htpb]
    \centering
    \includegraphics[width=0.9\textwidth]{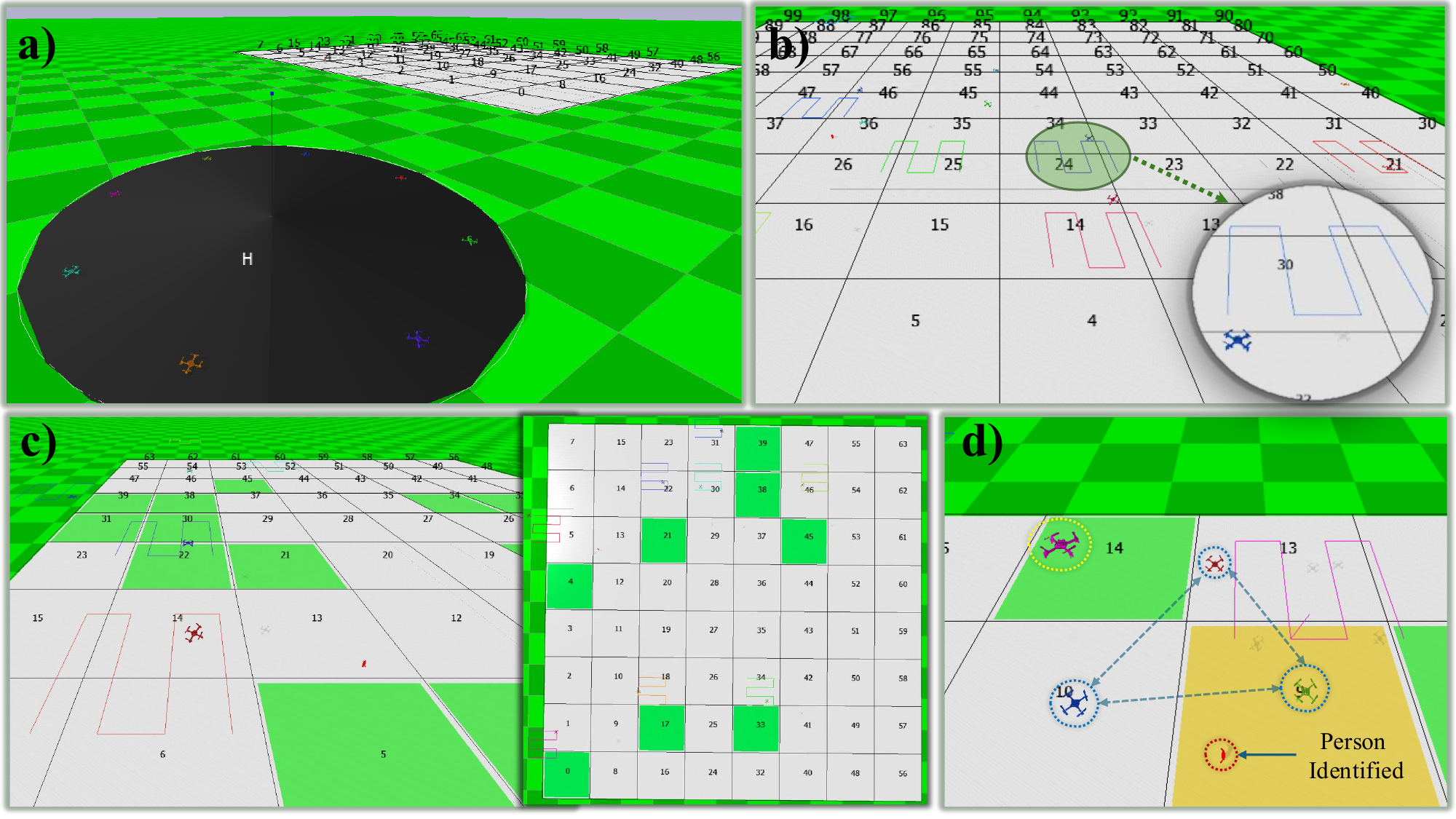} 
    \vspace{-3mm}
    \caption{Operational Sequence of the IRDS Architecture. \textbf{(a)} Swarm initialization and market bidding at the helipad. \textbf{(b)} Execution of local Boustrophedon coverage paths within assigned sectors. \textbf{(c)} Global state visualization where green cells denote cleared areas, validating the distributed memory. \textbf{(d)} Execution of the Geometric Consensus Protocol for mission: neighboring agents converge to verify a detected target (Person Identified), forming the required triangulation formation.}
    \label{fig:scenario_seq}
\end{figure*}

To evaluate the system's efficacy, we simulated a complete SAR mission. The operational sequence, seen in Fig. \ref{fig:scenario_seq}, proceeds through four distinct phases:

\begin{itemize}
    \item \textbf{Deployment (Fig. \ref{fig:scenario_seq}a):} The swarm initializes at a central helipad ($H$). During this phase, the \textit{Phase I} static market allocation distributes the initial set of target sectors to ensure uniform spatial distribution.
    
    \item \textbf{Distributed Search (Fig. \ref{fig:scenario_seq}b):} Upon reaching their assigned sectors, agents execute the \textit{Layer 2} local path planner. The inset highlights the generated trajectories within the assigned $1.5m^2$ grid cells.
    
    \item \textbf{State Synchronization (Fig. \ref{fig:scenario_seq}c):} As sectors are cleared, agents update the global belief state. Green cells indicate fully searched regions; white cells remain effectively "on the market." This visual feedback confirms the functionality of the decentralized ledger in preventing redundant searching.
    \item \textbf{Target Verification (Fig. \ref{fig:scenario_seq}d):} Upon detecting a potential victim (red dotted circle), the searching agent triggers the \textit{Layer 3} Consensus Protocol. The figure captures the "Geometric Lock" moment where neighboring agents (highlighted in yellow/blue dashed circles) converge to form the verification triangle, satisfying the unanimity condition required to confirm the target and conclude the mission.
\end{itemize}

\subsection{Experimental Setup}
The simulation is built upon the \texttt{gym-pybullet-drones} framework\footnote{Source code and documentation: \url{github.com/luizgiacomossi/Intelligent-Drone-Swarm/tree/paper}. Video of the simulator: \url{figshare.com/s/1b36e110c99e376c9f67}}. The search domain was defined as an $8 \times 8$ discretized grid ($64$ sectors total), where each sector represents a $1.5m \times 1.5m$ area. The target is randomly placed within the grid at the start of each episode.
\begin{itemize}
    \item \textbf{Agents:} Bitcraze Crazyflie 2.x dynamics.
    \item \textbf{Market Config:} Base Cost $C_{base} = 2.0$, Distance Factor $\delta = 10.0$.
    \item \textbf{Statistical Rigor:} Each configuration is executed for $K=100$ Monte Carlo trials for statistical significance given the stochastic nature of target placement.
\end{itemize}

\subsection{Performance Metrics}
Tab.\ref{tab:metrics} summarizes the Key Performance Indicators (KPIs) used to evaluate mission efficiency, system reliability, and market dynamics across all experiments.

\begin{table}[h]
    \centering
    \caption{Key Performance Indicators (KPIs) Definition}
    \label{tab:metrics}
    \renewcommand{\arraystretch}{1.25} 
    \resizebox{\columnwidth}{!}{%
        \begin{tabular}{|p{2.2cm}|p{7cm}|}
            \hline
            \textbf{Metric} & \textbf{Definition \& Significance} \\ 
            \hline
            \multicolumn{2}{|l|}{\textit{\textbf{1. Mission Efficiency}}} \\ 
            \hline
            Success Rate ($\eta$) & Percentage of trials where the target is identified and verified via consensus. \\ \hline
            Duration ($T_{total}$) & Time elapsed from mission start ($t_0$) until target verification. \\ \hline
            Distance ($D_{total}$) & The scalar sum of trajectories for all agents, serving as a direct proxy for total fleet energy consumption. \\ 
            \hline
            \multicolumn{2}{|l|}{\textit{\textbf{2. Reliability \& Resilience}}} \\ 
            \hline
            Reallocation Latency ($L_{realloc}$) & Time elapsed between an agent's failure and the market-based reassignment of its active tasks. \\ \hline
            Recovery Count ($C_{rec}$) & The absolute number of tasks transferred from faulty agents to healthy neighbors to heal coverage gaps. \\ 
            \hline
            \multicolumn{2}{|l|}{\textit{\textbf{3. Economic Dynamics}}} \\ 
            \hline
            Bid Density & Average number of valid bids per auction, indicating network connectivity and swarm availability. \\ \hline
            Gini Index ($G$) & Measures workload inequality ($0=$ balance, $1=$ monopoly); used to quantify the "Cost of Resilience" during load shedding. \\ 
            \hline
        \end{tabular}%
        
    }
    \vspace{-4mm}
\end{table}

\vspace{-1mm}

\subsection{Experimental Scenarios}
\subsubsection{Scenario A: Scalability Analysis}
We evaluate performance limits by increasing swarm density $N \in \{2, 4, 8\}$ (doubling resources) under nominal conditions. The objective is to quantify how quickly the swarm can locate a target as agent count increases.

\subsubsection{Scenario B: Fault Tolerance (Resilience)}
We fix the swarm size at $N=8$ and inject stochastic faults:
\begin{itemize}
    \item \textbf{Single Fault:} One agent triggers a \texttt{BAD\_BATTERY} fault at $t=8s$, forcing task release.
    \item \textbf{Double Fault:} Two agents fail in a sequence ($t=8s, t=15s$), reducing the workforce by $25\%$.
\end{itemize}

\vspace{-1mm}
\subsection{Quantitative Results}
\begin{figure*}[ht]
    \centering
    \includegraphics[width=0.9\textwidth]{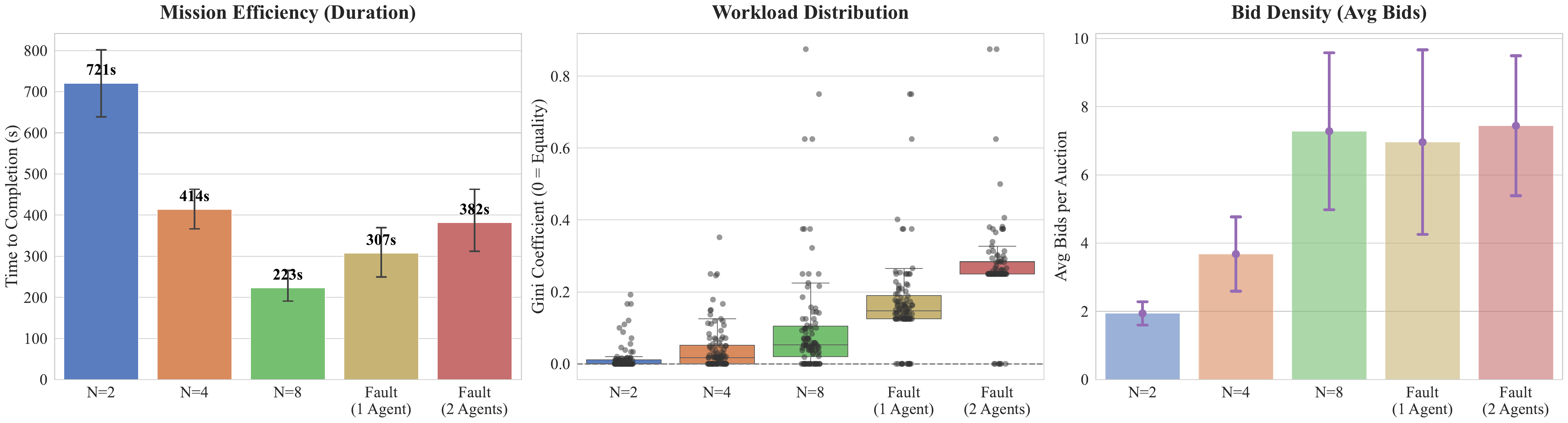} 
    \vspace{-3mm}
    \caption{\textbf{Exp. Results.} \textbf{(Left) Mission Efficiency:} Scalability is observed from $N=2$ to $N=8$. Error bars indicate the 95\% confidence interval, showing higher consistency at higher swarm densities. Fault scenarios show temporal degradation compared to nominal $N=8$ case, confirming the cost of agent loss. \textbf{(Center) Workload Distribution (Gini):} Inequality rises during faults as surviving agents absorb extra tasks. \textbf{(Right) Bid Density:} Bids per auction saturate at $N=8$, indicating high market availability, where nearly all agents actively participate in the allocation process.}  
    \label{fig:results}
\end{figure*}

The aggregated data from $K=100$ trials is presented in Table \ref{tab:comprehensive_results} and visualized in Fig. \ref{fig:results}.

\vspace{-2mm}
\begin{table}[h]
    \centering
    \caption{Comprehensive Performance Analysis (Mean $\pm$ Std. Dev.)}
    \label{tab:comprehensive_results}
    \setlength{\tabcolsep}{1.5pt} 
    \renewcommand{\arraystretch}{1.2} 
    \resizebox{\columnwidth}{!}{%
    \begin{tabular}{|l||c|c|c|c|c|c|c|}
    \hline
    \textbf{Scenario} & \textbf{Succ.} & \textbf{Time} & \textbf{Dist.} & \textbf{Recov.} & \textbf{Lat.} & \textbf{Bid} & \textbf{Gini} \\
     & \textbf{($\eta$)} & \textbf{($T_{total}$)} & \textbf{($D_{total}$)} & \textbf{($C_{rec}$)} & \textbf{($L_{realloc}$)} & \textbf{Dens.} & \textbf{($G$)} \\ \hline
    \hline
    \textbf{N=2 (Baseline)} & 97\% & $721 \pm 419$ & $\mathbf{346} \pm 187$ & - & - & $1.9 \pm 0.3$ & $0.02 \pm 0.04$ \\ \hline
    \textbf{N=4 (Nominal)} & \textbf{100\%} & $414 \pm 238$ & $400 \pm 195$ & - & - & $3.7 \pm 1.1$ & $0.04 \pm 0.06$ \\ \hline
    \textbf{N=8 (Saturated)} & 99\% & $\mathbf{223} \pm 190$ & $420 \pm 196$ & - & - & $7.3 \pm 2.3$ & $0.10 \pm 0.15$ \\ \hline
    \hline
    \textbf{Fault (1)} & 97\% & $307 \pm 308$ & $449 \pm 240$ & $6.5 \pm 10.1$ & $\mathbf{13.1} \pm 21.2$ & $7.0 \pm 2.7$ & $0.17 \pm 0.13$ \\ \hline
    \textbf{Fault (2)} & 93\% & $382 \pm 394$ & $453 \pm 228$ & $8.0 \pm 12.5$ & $17.7 \pm 26.6$ & $7.4 \pm 2.1$ & $0.27 \pm 0.13$ \\ \hline
    \end{tabular}%
    }
\end{table}

\subsubsection{Scalability and Efficiency}
Increasing the swarm size improves the time-to-find. As seen in Table \ref{tab:comprehensive_results}, the mean search duration drops from $721s$ ($N=2$) to $223s$ ($N=8$). Interestingly, the \textit{Total Distance} metric increases slightly with swarm size ($346m$ to $420m$), reflecting the overhead of coordinating a larger fleet; however, this energy cost is justified by the $3.2\times$ speedup in mission completion.

\subsubsection{Fault Tolerance Data}
In Scenario B, the system kept high success rate ($97.0\%$/$93.0\%$), with failures limited to motion deadlocks (local minima) rather than sensor errors. The \textit{Reallocation Latency} was measured at $13.1s$ and $17.7s$, respectively. Crucially, the \textit{Recovery Count} ($C_{rec} \approx 6.5$) shows that if an agent fails, its sectors are immediately re-auctioned and picked up by neighbors, preventing "blind spots" where the target might be.

\section{Discussion}

\label{sec:discussion}

The results supports the core hypothesis of IRDS: a decentralized and cost-minimizing market can serve as a robust proxy for complex centralized replanning in safety-critical environments. In this section, we analyze the emergent behaviors of the swarm and the trade-offs inherent in our proposed architecture.

\subsection{Operational Resilience and Latency}
A key metric for safety-critical SAR is the system's reaction speed to failure. As seen in Tab.\ref{tab:comprehensive_results}, the average Reallocation Latency ($L_{realloc}$) of roughly $13-18s$ represents approximately $4.3$--$4.6\%$ of the total mission duration ($307s$ and $382s$, respectively), and is small relative to the mission's operational timescale. This speed is a consequence of the \textit{Liquidation Protocol} defined in Sec. \ref{sec:method}. In traditional leader-election algorithms (e.g., Raft), a node failure triggers a timeout and a voting phase, consuming valuable time. In IRDS, failure is treated instantaneously as a market event: the "goods" (sectors) are released, and the "buyers" (neighbors) immediately acquire them because their proximity makes them the lowest bidders. 

\vspace{-2mm}
\subsection{Emergent Economic Dynamics}

An analysis of the metrics shows how the virtual economy regulates swarm behavior without explicit control.

\subsubsection{Market Availability}
The \textit{Bids per Auction} metric (Table \ref{tab:comprehensive_results}) indicates high system connectivity. At $N=8$, participation saturates at $\approx 7.3$ bids per auction ($91\%$ of the swarm). This "High Availability" ensures that task allocation is globally optimized rather than locally greedy, as nearly every agent is aware of every opportunity. Unlike limited-range heuristics that suffer from information silos, the auction mechanism maintains global awareness without saturating the decision-making process.

\subsubsection{The Cost of Resilience (Gini Index)}
The Gini Coefficient highlights the unavoidable trade-off between fairness and resilience. In the baseline ($N=8$, No Fault), the Gini index is low ($0.10$), indicating that the market equalizes travel distance across the fleet. However, under the "Double Fault" scenario, the Gini index rises to $0.27$. This inequality represents the calculated \textit{Cost of Resilience}: surviving agents near the failure site must undertake disproportionate work to cover the gap. Crucially, the system stabilizes at $G=0.27$ rather than diverging to $G=1.0$ (monopolization). This proves that the budget constraint $W_i(t)$ effectively prevents any single agent from depleting its battery to save the mission, enforcing a soft load-balancing even during crisis management.

\subsection{Scalability vs. Energy Overhead}
While the system achieves linear speedup from $N=2$ ($721s$) to $N=8$ ($223s$), the \textit{Total Distance} metric reveals a hidden cost. The collective distance traveled increases from $346m$ to $420m$ as the swarm grows. This $21\%$ increase in energy expenditure represents the overhead of coordination and initial deployment (agents traveling from the depot to distant sectors). For safety-critical SAR, this is an acceptable trade-off: the priority is minimizing time-to-rescue, even at the cost of higher  energy consumption.

\subsection{Consensus Latency and Physical Deadlocks}
While the Dynamic Quorum mechanism handles agent crashes (Health Deadlocks), the current implementation does not enforce a timeout for the voting. This exposes the system to \textit{Motion Deadlocks}, where a healthy verifier becomes trapped by obstacles or other agents and cannot reach its voting position. This phenomenon explains the high variance observed in $N=8$ baseline ($\sigma = 190s$) and Double Fault ($\sigma = 394s$) scenarios. In crowded environments, the APF repulsion forces can create local minima that trap verifiers, causing the consensus protocol to stall. Future iterations may mitigate this by implementing a $T_{max}$ timeout, causing the system to abort the vote if physical convergence is not achieved within a bounded window.

\subsection{Algorithmic Complexity Analysis}
An advantage of IRDS is its computational scalability compared to global optimization methods.
\begin{itemize}
    \item \textbf{Centralized Global Optimization (Hungarian):} Solving the linear assignment problem for $N$ agents and $M$ tasks usually requires $\mathcal{O}(N^3)$ time. While feasible for small teams, this polynomial growth creates a computational bottleneck as the swarm size increases.
    \item \textbf{IRDS Greedy Reverse Auction:} Each agent calculates $M$ bids locally with complexity $\mathcal{O}(M)$. The allocation logic effectively sorts the bids to find the lowest cost. Using a standard efficient sort (e.g., Mergesort), this operation scales as $\mathcal{O}(K \log K)$, where $K = N \times M$ is the total number of bids.
\end{itemize}
Thus, the IRDS complexity scales quasi-linearly: $\mathcal{O}(NM \log (NM))$. This lower complexity class ensures that the architecture remains viable for large-scale deployments where $\mathcal{O}(N^3)$ methods would become computationally prohibitive.

\section{Conclusion and Future Work}
\label{sec:conclusion}
This work presented the \textit{Intelligent Replanning Drone Swarm} (IRDS), a coordination architecture for safety-critical Search and Rescue. By replacing static planning with a dynamic \textit{Reverse-Auction Market}, we transformed the problem of agent failure into a mechanism for economic opportunity, where the loss of a node automatically triggers a localized cost-minimized task reallocation.

The architecture was tested using physics-based simulations ($K=100$ trials). Results indicate that IRDS is \textit{Fault-Tolerant}: the swarm successfully recovers tasks from failed agents with an average reaction latency of $13.1s$ (single fault). Furthermore, the system maintained a mission success rate of $93\%$ even under $25\%$ agent loss, supporting its viability for fault-prone environments. Scaling the swarm from $N=2$ to $N=8$ yielded a $3.2\times$ speedup in time-to-completion at the cost of a modest increase in total energy expenditure due to coordination overhead. Future work includes deployment on Bitcraze Crazyflie drones to verify communication overhead under realistic RF packet loss, extending the market model to heterogeneous swarms with sensor-aware bidding, integrating Bayesian search probability models, and evaluating scalability limits with larger swarms ($N=16, 32$).


\section*{acknowledgment}
\vspace{-1mm}
This research was funded by the European Union’s Horizon Europe research and innovation programme and the Chips Joint Undertaking under Grant Agreement No. 101194287, project NexTArc (Next Generation Open Innovations in Trustworthy Embedded AI Architectures for Smart Cities, Mobility and Logistics).


\bibliographystyle{IEEEtran}
\bibliography{references}

\end{document}